\documentclass[Afour,sageh,times]{sagej}

\usepackage{moreverb,url}
\usepackage{xcolor}
\usepackage[colorlinks,bookmarksopen,bookmarksnumbered,citecolor=blue,linkcolor=blue,urlcolor=blue]{hyperref}
\newcommand{\doi}[1]{{\urlstyle{rm}DOI:\nobreakspace\href{https://doi.org/#1}{\nolinkurl{#1}}}}
\usepackage{amsmath,amsfonts,amssymb,bm}
\allowdisplaybreaks[2]
\usepackage{booktabs}
\usepackage{multirow}
\usepackage{placeins}
\usepackage{siunitx}
\usepackage[ruled,vlined,linesnumbered]{algorithm2e}
\definecolor{mdobstatebox}{RGB}{219,238,243}
\definecolor{mdobcostbox}{RGB}{239,223,209}
\definecolor{mdobfeedback}{RGB}{151,57,46}
\definecolor{mdobcontrol}{RGB}{88,111,30}

\newcommand{\vect}[1]{\bm{#1}}
\newcommand{\state}{\vect{x}}
\newcommand{\query}{\vect{q}}

\newcommand{\action}{\vect{u}}
\newcommand{\dist}{\vect{d}}

\newcommand{\env}{\mathcal{E}}
\newcommand{\pcost}{c}

\newcommand{\distspace}{\mathcal{D}}
\newcommand{\queryspace}{\mathcal{Q}}

\newcommand{\fb}{\mathrm{fb}}
\newcommand{\figref}[1]{\hyperref[#1]{Figure.~\ref*{#1}}}
\DeclareMathOperator*{\argmin}{arg\,min}
\theoremstyle{definition}
\newtheorem{remark}{Remark}
\theoremstyle{sage}

\makeatletter
\renewcommand{\@maketitle}{%
  \begin{center}
    {\sffamily\titlesize\bfseries\@title\par}
    \vspace{12pt}
    {\sffamily\large\bfseries\@author\par}
  \end{center}
  \vspace{10pt}
  \noindent\usebox\absbox\par
  \vspace{8pt}
  {\sffamily\noindent\@keywords\par}
  \vspace{18pt}
}
\let\ps@title\ps@plain
\makeatother
\begin{document}

\runninghead{}

\title{Tackling Sim-to-Real Mismatch Through Sampling-Based Disturbance Observers: From Analytical Models to Learned World Models}

\author{Tianqi Zhu\affilnum{1}, Jun Yang\affilnum{2}, Jianliang Mao\affilnum{3}, Cong Li\affilnum{2}, and Shihua Li\affilnum{1}}

\affiliation{%
\affilnum{1}School of Automation, Key Laboratory of Measurement and Control of CSE, Ministry of Education, Southeast University, Nanjing 210096, China\\
\affilnum{2}Robotics and Autonomous Systems Thrust, Systems Hub, The Hong Kong University of Science and Technology (Guangzhou), Guangzhou 511453, China\\
\affilnum{3}Faculty of Artificial Intelligence, Shanghai University of Electric Power, Shanghai 200090, China}

\corrauth{Jun Yang, The Hong Kong University of Science and Technology (Guangzhou), Guangzhou 511453, China; Jianliang Mao, Shanghai University of Electric Power, Shanghai 200090, China.}

\email{jy@hkust-gz.edu.cn (Jun Yang); jl\_mao@shiep.edu.cn (Jianliang Mao)}

\begin{abstract}
Robotic controllers increasingly rely on analytical models, simulators, cost-query interfaces, and learned world models. However, physical deployment can deviate from nominal assumptions, and additional disturbances may arise even when the model itself is accurate. In control systems, disturbance observers (DOB) are widely used to estimate such unmeasured effects from nominal models and measured feedback. Classical DOB formulations are generally built around explicit plant models. This paper develops the sampling-based disturbance observer (SDOB), extending the DOB principle to a broader range of models, including simulators and learned world models, through state-rollout or cost-query interfaces. SDOB separates two observable channels: state-effect disturbances, for which the feedback state differs from its prediction, and cost disturbances, for which the same query state receives different costs as the perceived environment changes. Diverse simulation and real-robot experiments across traditional and learned models demonstrate the effectiveness of SDOB in compensating for sim-to-real mismatch and improving control performance.
\end{abstract}


\maketitle

\section{Introduction}
From early model-based control to modern robotics built on simulators and learned models, deployment has always involved discrepancies between simulator models and physical plant. In classical systems, these discrepancies appear as parameter uncertainty, unmodeled dynamics, and external disturbances \citep{ohnishi1996_dob,chen2016_nldo}; in contemporary robotics, the same problem is often discussed as the sim-to-real gap \citep{openai2020_dexterous_sim2real,jaquier2025_transfer_learning_review}. Such discrepancies can degrade tracking performance and closed-loop stability. If a disturbance is measurable, feedforward compensation can often reduce its influence. In practice, the relevant disturbances are often unmeasured or entangled with model uncertainty. Classical DOB formulations estimate these effects in systems described by explicit plant or low-dimensional state-space models \citep{ohnishi1996_dob,chen2016_nldo,sariyildiz2020_dob_survey}. Conventional MPC operates within the same model-based setting, optimizing behavior over a finite horizon using explicit dynamics and constraints \citep{garcia1989_mpc_survey,mayne2000_constrained_mpc,ostafew2016_nmpc}. DOB estimates can therefore be incorporated into its prediction and optimization loop, for example by correcting output predictions \citep{yang2015_dob_mpc}.

However, modern robotic systems increasingly use sampling-based MPC when the dynamics or costs are highly nonlinear, nonsmooth, or accessible only through simulators. Sampling-based MPC evaluates candidate action sequences through forward rollouts. Methods such as model predictive path integral control (MPPI) \citep{williams2017_mppi_tro,carius2022_path_integral,koptev2024_sampling_mpc} and the cross-entropy method (CEM) \citep{rubinstein1997_cross_entropy,kobilarov2012_cross_entropy_planning,pinneri2020_cem} rank these candidates using scalar trajectory costs. Because they require only forward rollouts and scalar costs, these methods can use nonlinear simulators without analytic derivatives. At the same time, rollout evaluation has expanded beyond analytical dynamics to perception-derived cost maps and signed-distance fields \citep{octomap2013,oleynikova2017_voxblox,zucker2013_chomp}, while world models \citep{hafner2019_planet,hafner2020_dreamer} learn action-conditioned latent dynamics for prediction and planning. This flexibility allows MPC to use a broader range of models, but it also makes disturbance estimation less direct. The resulting interfaces may provide neither an invertible plant representation nor a clearly defined input-equivalent disturbance channel. Unmodeled effects are therefore revealed through deviations in the feedback state or changes in the costs assigned to queried states, for example after contact, perception updates, or obstacle motion. The controller observes these effects only after executing an action and receiving feedback. Taken together, these changes leave a mismatch between the information required by classical DOB formulations and the interfaces exposed by modern predictive models.

To address this interface mismatch, we present SDOB as a framework for feedback-based disturbance compensation in analytical models, simulators, and learned predictors exposed through rollout or cost-query interfaces. We refer to each member of this family as a modern model. The state channel fits feedback transitions within a task-defined disturbance parameterization, while the cost channel models local transport of environment-dependent costs. Together, these channels specify how observed mismatch enters subsequent rollout and cost evaluation. SDOB therefore targets locally persistent, task-relevant mismatch rather than complete identification of the sim-to-real gap. The main contributions are:
\begin{itemize}
\item[{1)}\hspace{0.5em}]
A new interface-level disturbance representation is introduced for modern models. Unlike conventional DOB formulations that primarily describe disturbances through plant dynamics and state evolution, SDOB separates observable deployment mismatch into two complementary channels: a state-effect disturbance that alters rollout states and a cost disturbance that changes costs at fixed query states. This extends disturbance observation to environment-dependent cost evaluation.
\item[{2)}\hspace{0.5em}]
Based on this representation, SDOB introduces a next-generation approach to disturbance observation that operates through forward rollouts and cost queries. Unlike classical DOB designs built on explicit transfer functions or state-space equations, SDOB does not require access to the analytical expressions of the underlying model. It estimates both disturbance channels online and incorporates their effects into future state propagation and cost evaluation, without plant inversion or model retraining.
\item[{3)}\hspace{0.5em}]
The framework is extensively validated across diverse simulation environments and real-robot platforms, spanning analytical-model disturbance compensation, moving-obstacle navigation, latent and visual world-model planning, mobile-robot navigation, dual-arm obstacle avoidance, and humanoid dual-arm manipulation with PointWorld, demonstrating its applicability across different model classes and robotic tasks.
\end{itemize}

The remainder of this article is organized as follows. Section 2 reviews related work. Section 3 introduces the classical DOB formulation used as a reference. Section 4 presents SDOB, including its deployment-oriented interpretation, two disturbance channels, and compensation interfaces. Section 5 reports the simulation and real-robot experiments. Section 6 concludes the article and discusses its limitations and future directions.

\section{Related Work}

SDOB connects disturbance observation, sampling-based MPC, dynamic environment estimation and cost prediction, sim-to-real adaptation, and world models. Disturbance observation provides the observer principle, sampling-based MPC provides rollout and cost evaluation interfaces, and sim-to-real adaptation motivates correction under deployment mismatch. Dynamic environment estimation and cost prediction provide the context for the cost disturbance channel. World models represent a frontier class within the modern models defined above and provide learned, action-conditioned representations for predicting future states.

\subsection{DOB-Based Control}
Classical DOB-based control estimates input disturbances from a nominal plant, commonly through an inverse model and low pass filtering, to improve motion control robustness \citep{ohnishi1996_dob,shim2009_dob_stability,sariyildiz2020_dob_survey,park2024_dob_survey}. Nonlinear DOB formulations extend this principle to robotic dynamical systems \citep{chen2000_nldo_robot,chen2000_design_nldo,mohammadi2013_nldo_nonlinear,chen2016_nldo}. Active disturbance rejection control (ADRC) and extended state observer (ESO) methods likewise estimate lumped uncertainty in feedback loops \citep{han2009_adrc,gao2003_adrc_tuning}, while DOB-enhanced predictive control incorporates disturbance estimates into prediction and optimization \citep{yang2015_dob_mpc}. Many classical DOB formulations assume an explicit plant model and a predefined disturbance channel. These assumptions are difficult to retain when control operates on images, latent or object centric states, or rollout distributions available only through forward simulation and cost evaluation.

\subsection{Sampling-Based MPC}
Sampling-based MPC optimizes candidate action sequences through forward rollouts and ranks, weights, or refits them using trajectory costs \citep{theodorou2010_path_integral,williams2016_mppi,williams2017_mppi_tro,pinneri2020_cem}. Requiring only simulated futures and scalar costs makes it suitable for nonlinear, nonsmooth, and black box models. Tube-MPPI uses MPPI to generate a nominal trajectory and an ancillary tracking controller to keep the actual state within a tube around that trajectory \citep{williams2018_tube_mppi}. Robust model predictive path integral control (RMPPI) extends this architecture with augmented rollouts of the nominal and actual dynamics, tracking-aware importance sampling, and safety logic for nominal-state propagation \citep{gandhi2021_mppi_quad}. Parameter uncertainty can be incorporated directly into path integral control \citep{abraham2020_mppi_covariance}, while Shield-MPPI integrates control barrier functions for safety \citep{yin2023_shield_mppi}.

\subsection{Dynamic Environment Estimation and Cost Prediction}
Planning in dynamic environments requires anticipating how environmental motion changes collision risks and trajectory costs. Dynamic occupancy grids estimate occupancy and motion \citep{nuss2018_dynamic_occupancy}, while Occupancy Flow Fields jointly predict their future evolution \citep{mahjourian2022_occupancy_flow}. ImplicitO supports continuous occupancy and flow queries at selected locations, avoiding dense scene prediction \citep{agro2023_implicit_occupancy_flow}. To evaluate candidate trajectories, neural motion planners learn spatiotemporal cost volumes \citep{zeng2019_neural_motion_planner}, while QuAD queries predicted occupancy at planning-relevant locations \citep{biswas2024_quad}. These methods primarily focus on predicting occupancy or planning costs, with less emphasis on treating cost changes at fixed query states as a distinct disturbance channel for online estimation and compensation.

\subsection{Sim-to-Real Adaptation}
Sim-to-real adaptation takes a complementary approach by changing the training distribution, system model, or simulator. Domain randomization broadens the training variation \citep{tobin2017_domain_randomization}, system identification calibrates the dynamics \citep{tan2018_agile_locomotion}, and feedback driven simulator adaptation updates the simulator from real experience \citep{chebotar2019_closing_sim2real_loop}. Adaptation methods can align latent representations under unknown target appearances \citep{yoneda2022_latent_alignment}, while ManiDreams propagates perceptual, parametric, and structural uncertainty through task specific predictive models \citep{wang2026_manidreams}. Such methods adapt the simulator, representation, policy, or predictive state, but remain tied to variations represented during training or adaptation. Greater simulator fidelity also does not guarantee that method rankings or task outcomes transfer to hardware \citep{kadian2020_sim2real_predictivity}, particularly when learned states do not map directly to physical variables.

\subsection{World Models for Prediction and Planning}
World models learn action conditioned representations for imagined futures, from early latent models \citep{ha2018_world_models} to PlaNet \citep{hafner2019_planet}, Dreamer \citep{hafner2020_dreamer,hafner2025_mastering}, and TD-MPC2 \citep{hansen2024_tdmpc2}. Learned dynamics for manipulation \citep{ai2025_learning_based_dynamics_review}, broader predictive architectures \citep{lecun2022_path_ami}, V-JEPA \citep{bardes2024_vjepa}, and PointWorld \citep{huang2026_pointworld} extend this direction toward feature level and 3D action conditioned representations. These models improve prediction, but nominal replanning does not usually preserve recurring execution mismatch as an explicit disturbance. Actuator bias, contact mismatch, object drift, and environmental change can therefore reappear as rollout error at every control cycle.

These developments motivate extending disturbance observation to a broader range of predictive models. To establish the conceptual basis for this extension, the next section revisits the basic principle of classical DOBs.

\section{Preliminaries}

\begin{figure}[!t]
\centering
\includegraphics[width=0.98\columnwidth]{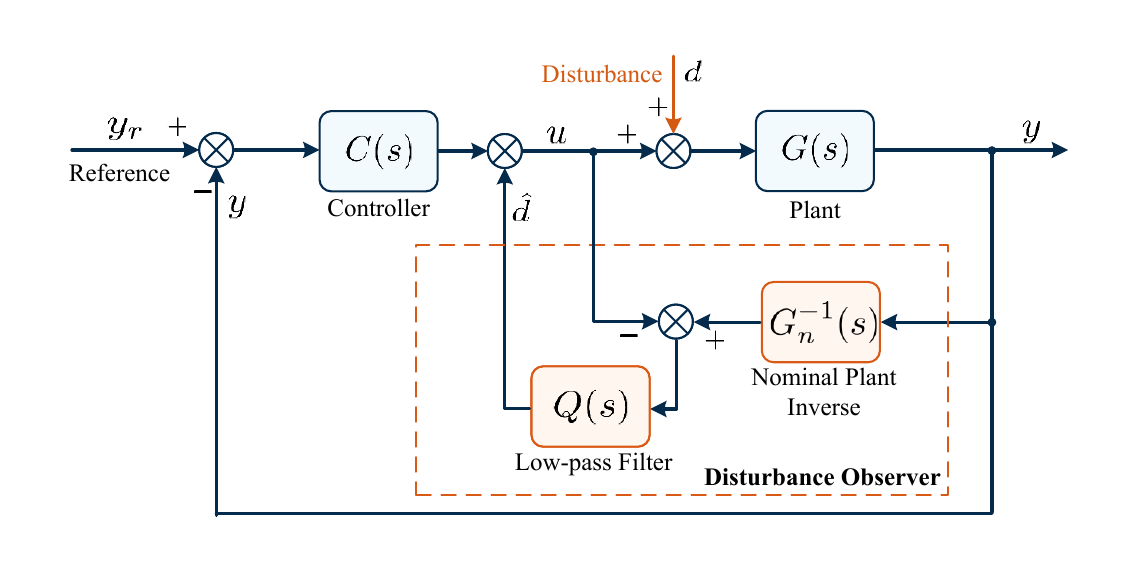}
\caption{Classical frequency-domain DOB control structure.}
\label{fig:classical_dob_structure}
\end{figure}

\begin{figure*}[!t]
\centering
\includegraphics[width=0.98\textwidth]{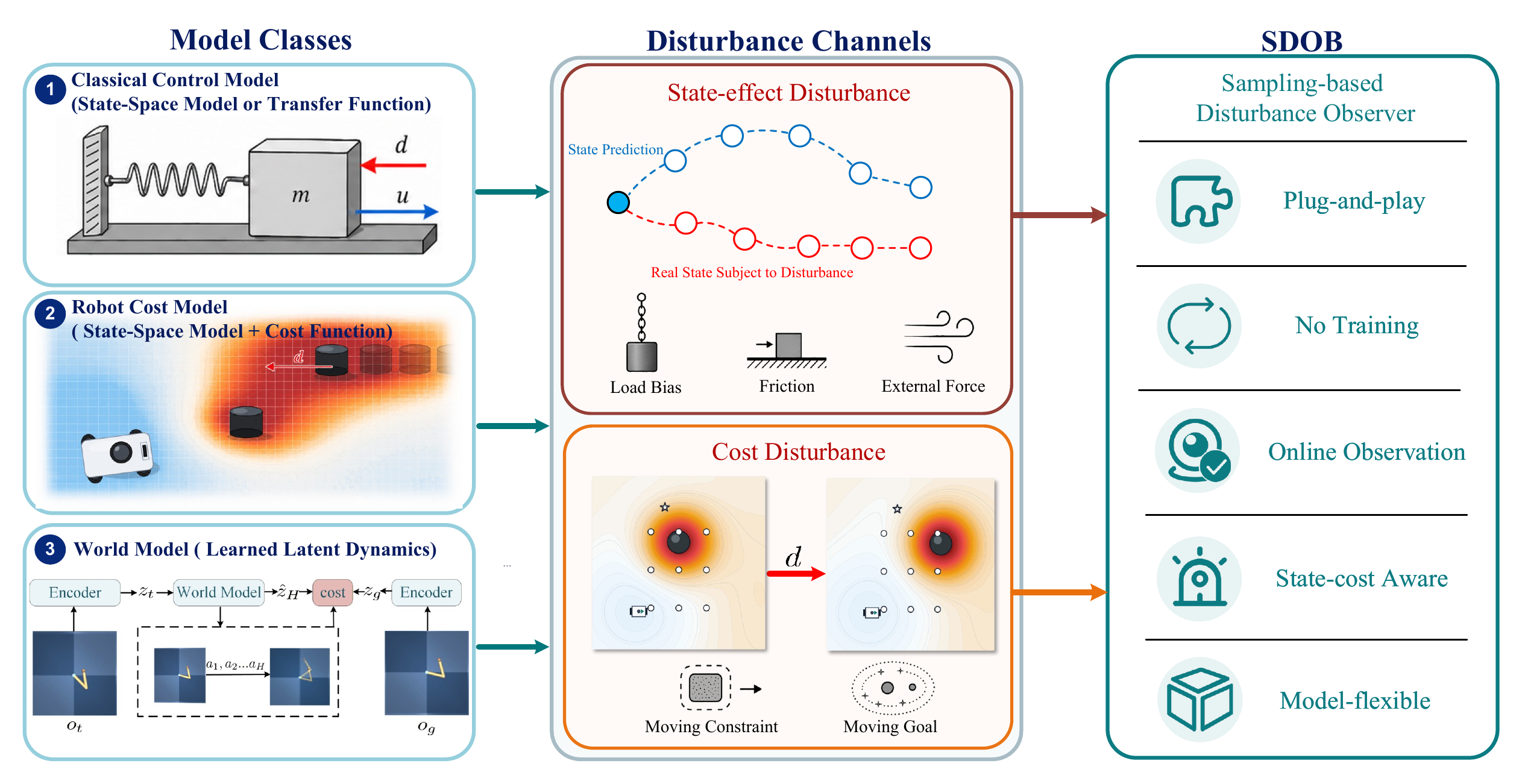}
\caption{Overview of SDOB: model classes, disturbance channels, and key advantages.}
\label{fig:mdob_overview}
\end{figure*}

\subsection{Disturbance Observer}
\figref{fig:classical_dob_structure} shows the classical frequency-domain DOB structure. For clarity, consider a single-input single-output plant with a
matched input-side disturbance:
\begin{equation}
  y(s)=G(s)\left(u(s)+d(s)\right)
  \label{eq:generic_disturbed_plant}
\end{equation}
where $G(s)$ is the actual physical plant, $u(s)$ is the applied control input,
and $d(s)$ is an unknown disturbance entering through the same channel as the
input. In this formulation, $G_n(s)$ denotes the nominal plant model used for disturbance estimation. The nominal inverse
$G_n^{-1}(s)$ maps the measured output back to an input-equivalent quantity,
which is then compared with the applied input and filtered before compensation
\citep{ohnishi1996_dob,chen2016_nldo,sariyildiz2020_dob_survey}.

The DOB estimate is
\begin{equation}
  \hat d(s)=Q(s)\left(G_n^{-1}(s)y(s)-u(s)\right),
  \label{eq:generic_dob}
\end{equation}
where $Q(s)$ is a stable low-pass filter. The estimate is then used in the
input channel as
\begin{equation}
  u(s)=u_c(s)-\hat d(s),
  \label{eq:generic_dob_compensation}
\end{equation}
where $u_c(s)$ is the baseline controller output before disturbance
compensation.

\paragraph{Example: discrete-time integrator.}
A minimal discrete-time example is a scalar integrator with a matched
disturbance,
\begin{equation}
  y_{k+1}=y_k+u_k+d_k,
  \label{eq:integrator_disturbed}
\end{equation}
with nominal prediction
\begin{equation}
  \hat y_{k+1|k}=y_k+u_k .
  \label{eq:discrete_integrator_prediction}
\end{equation}
After executing $u_k$ and observing $y_{k+1}$, the one-step prediction error is
\begin{equation}
  e_{k+1}=y_{k+1}-\hat y_{k+1|k}=y_{k+1}-y_k-u_k .
  \label{eq:integrator_inverse_residual}
\end{equation}
If the nominal model is exact and measurement noise is ignored, then
$e_{k+1}=d_k$. In practice, this residual is passed through a discrete
low-pass filter, for example
\begin{equation}
  \hat d_k=(1-\alpha)\hat d_{k-1}+\alpha e_{k+1},
  \qquad
  0<\alpha\leq 1,
  \label{eq:integrator_dob}
\end{equation}
and the next input can be compensated using the previous disturbance estimate,
\begin{equation}
  u_k=u_{c,k}-\hat d_{k-1}.
  \label{eq:integrator_compensation}
\end{equation}
The observer thus uses the filtered prediction error to compensate for the disturbance in the next control step.

\section{Sampling-Based Disturbance Observer}

As shown in \figref{fig:mdob_overview}, SDOB extends the classical DOB principle to models accessed through forward rollouts or cost queries. It uses the mismatch between a model query and the following feedback to estimate a correction for the next control step.

\subsection{Model and Forward Equations}
Let $\mathcal X$ denote the known state space, $\mathcal U$ the input space, $\queryspace$ the cost-query space, $\env$ the perceived-environment space, and $\distspace$ the disturbance space. At the beginning of step $k$, the controller has the latest feedback state $\state_k^{\fb}$, the perceived environment $\env_k$, and the cost-query function $\pcost_k$. Once the feedback state is passed to the controller, we write $\state_k\triangleq\state_k^{\fb}$. For a query state $\query\in\queryspace$,
\begin{equation}
  \pcost_k(\query)
  \triangleq
  \pcost(\query,\env_k).
  \label{eq:queried_cost_distribution}
\end{equation}
Here, $\pcost_k(\query)$ evaluates the query state $\query$ in the current perceived environment $\env_k$. The query state collects the information needed for cost evaluation, such as a vehicle pose or joint configuration for collision checking. It is obtained from the known state $\state$ through a task-specific mapping:
\begin{equation}
  \query
  =
  \Pi_q(\state),
  \qquad
  \Pi_q:\mathcal X\rightarrow\queryspace .
  \label{eq:query_state_extraction}
\end{equation}

\begin{figure}[t]
\centering
\begin{minipage}{0.98\columnwidth}
\centering
\includegraphics[width=\linewidth]{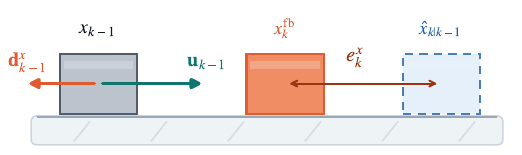}\\[-0.2em]
{\footnotesize\textbf{(a)} State-effect disturbance}
\end{minipage}
\vspace{0.45em}

\begin{minipage}{0.98\columnwidth}
\centering
\includegraphics[width=\linewidth]{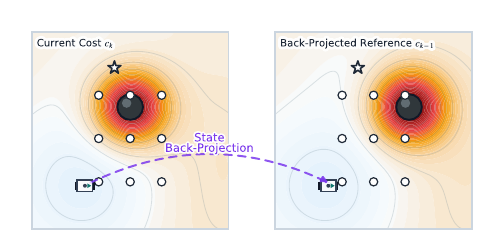}\\[-0.2em]
{\footnotesize\textbf{(b)} Cost disturbance}
\end{minipage}
\caption{Two disturbance channels used by SDOB. (a) Illustration under the state-effect disturbance $\dist_{k-1}^x$. (b) Illustration under the cost disturbance $\dist_{k-1}^c$.}
\label{fig:mdob_disturbance_channels}
\end{figure}

The nominal transition model is $F_x:\mathcal X\times\mathcal U\rightarrow\mathcal X$. The task also supplies a disturbance-enabled rollout interface
\begin{equation}
  F_x^d:
  \mathcal X\times\mathcal U\times\distspace
  \rightarrow
  \mathcal X,
  \qquad
  F_x^d(\state,\action,\vect{0})=F_x(\state,\action).
  \label{eq:state_disturbance_interface}
\end{equation}
The choice of $F_x^d$ determines the disturbance effects that the state channel can estimate. For example, $F_x^d$ may incorporate an input bias or an additive residual in the predicted state. For each task, we first specify how the disturbance affects the model prediction based on the available knowledge of the system. The interface $F_x^d$ then allows candidate disturbance values to be evaluated through forward prediction. SDOB compares the predicted transitions with the observed feedback to estimate a disturbance correction for subsequent rollouts. The resulting estimate is therefore a control-equivalent state effect, not necessarily a measured force or a unique physical parameter.

At the beginning of step $k$, the most recently completed transition is from $k-1$ to $k$. Its nominal state prediction and nominal cost-query function are
\begin{align}
  \hat{\state}_{k|k-1}
  &=
  F_x(\state_{k-1},\action_{k-1}),
  \label{eq:nominal_known_environment}\\
  \hat{\pcost}_{k|k-1}(\query)
  &=
  \pcost_{k-1}(\query)
  =
  \pcost(\query,\env_{k-1}).
  \label{eq:nominal_cost_function}
\end{align}

After $\action_{k-1}$ is executed, the feedback process returns the current state $\state_k^{\fb}$ and perceived environment $\env_k$. With an unknown state-effect disturbance, the completed transition and the current cost query can be written as
\begin{equation}
  \begin{aligned}
  \state_k^{\fb}
  &=
  F_x^d(\state_{k-1},\action_{k-1},\dist_{k-1}^x),
  \\
  \pcost_k(\query)
  &=
  \pcost(\query,\env_k) .
  \end{aligned}
  \label{eq:actual_full_environment}
\end{equation}
Thus $\hat{\state}_{k|k-1}$ and $\hat{\pcost}_{k|k-1}$ are formed before execution, whereas $\state_k^{\fb}$, $\env_k$, and $\pcost_k(\query)$ become available afterward. Here $\dist_{k-1}^x$ denotes the unknown state-side effect acting over the completed interval.

\subsection{Two Disturbance Channels}
SDOB separates mismatch by where it appears. A state-effect disturbance changes the feedback state, for example through load, friction, contact, or actuator bias. A cost disturbance changes the cost assigned to the same query state as the perceived environment changes. This second channel is useful when a point cloud, occupancy layer, or collision model is available for queries but is not maintained as a compact dynamic state. As illustrated in \figref{fig:mdob_disturbance_channels},
\begin{equation}
  \dist_{k-1}=(\dist_{k-1}^x,\dist_{k-1}^c(\cdot))
  \label{eq:disturbance_components}
\end{equation}
where $\dist_{k-1}^x$ denotes a state-effect disturbance and $\dist_{k-1}^c(\query)$ denotes the cost disturbance at query state $\query$. In this work, $\dist_{k-1}^c(\cdot)$ is a query-dependent vector field that describes the apparent local transport rate of the queried cost, rather than the physical velocity of an individual object.

\begin{remark}
SDOB requires three conditions. First, the controller must provide a repeatable rollout interface or comparable cost queries across consecutive updates. Second, the selected disturbance coordinates must affect the available feedback or queried costs; components that do not affect these signals are unobservable. Third, the disturbances are assumed to vary slowly over successive control updates, allowing their current estimates to be used for short-horizon prediction. The state channel additionally requires the disturbance-enabled interface in Eq.~\eqref{eq:state_disturbance_interface}. The cost channel assumes that local cost changes over one update can be approximated by short-horizon transport. Abrupt topology changes or several unresolved motions in the same neighborhood can violate this assumption.
\end{remark}

\begin{figure}[t]
\centering
\includegraphics[width=\columnwidth]{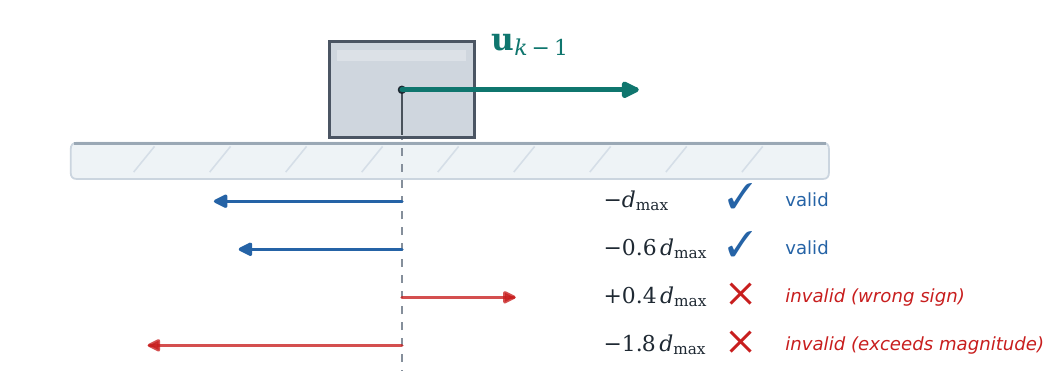}
\caption{Physical admissibility of sampled state-effect disturbances.}
\label{fig:mdob_three_term_system}
\end{figure}

\begin{figure*}[t]
\centering
\includegraphics[width=0.98\textwidth]{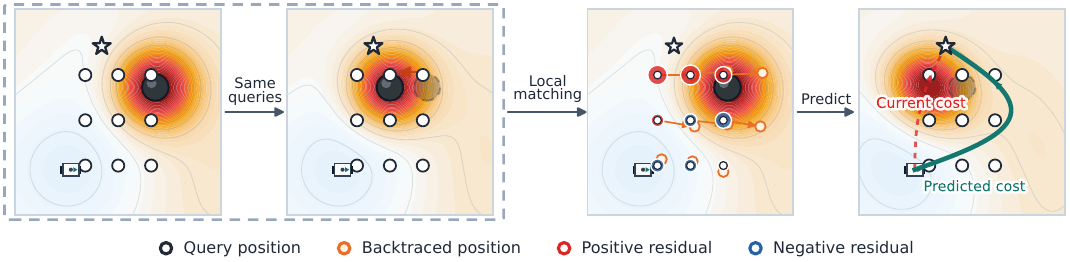}
\caption{Cost disturbance estimation and planning compensation. The first two panels query costs at the same locations before and after an environment update. The third panel matches each current query to a position in the previous field with similar surrounding costs. Arrows connect the current queries, marked by white centers, to these matched earlier positions, marked by orange circles. The displacement between them estimates how the local cost pattern has moved. Red rings indicate that the current cost is higher than the matched previous cost, while blue rings indicate that it is lower. The matching minimizes the neighborhood loss in Eq.~\eqref{eq:cost_disturbance_argmin}. The final panel uses the estimated motion to predict future costs and guide planning through Eqs.~\eqref{eq:cost_query_backtrace} and \eqref{eq:cost_query_transport}.}
\label{fig:cost_flow_estimation_pipeline}
\end{figure*}

\subsection{State-Effect Disturbance Estimation}
At the beginning of step $k$, the controller has received $\state_k^{\fb}$ after executing $\action_{k-1}$. Let $\hat{\dist}_{k-1}^x$ denote a candidate estimate of the unknown disturbance $\dist_{k-1}^x$ that acted during this completed transition. The candidate estimate produces the prediction
\begin{equation}
  \hat{\state}_{k|k-1}(\hat{\dist}_{k-1}^x)
  =
  F_x^d(\state_{k-1},\action_{k-1},\hat{\dist}_{k-1}^x).
  \label{eq:state_effect_nominal_prediction}
\end{equation}
The corresponding state-effect residual is the difference between the feedback state and this prediction
\begin{equation}
  \vect{e}_{k}^{x}(\hat{\dist}_{k-1}^x)
  =
  \state_{k}^{\fb}
  -
  \hat{\state}_{k|k-1}(\hat{\dist}_{k-1}^x).
  \label{eq:state_effect_disturbance}
\end{equation}
The residual in Eq.~\eqref{eq:state_effect_disturbance} scores how well each candidate disturbance explains the completed feedback transition. The disturbance estimate is obtained by minimizing
\begin{align}
  \hat{\dist}_{k-1}^{x,\star}
  &\in
  \argmin_{\hat{\dist}_{k-1}^x\in\distspace}
  \mathcal L_{k-1}^x(\hat{\dist}_{k-1}^x),
  \label{eq:state_effect_argmin}\\
  \mathcal L_{k-1}^x(\hat{\dist}_{k-1}^x)
  &=
  \underbrace{
  \ell_{\mathrm{pred},k-1}^x(\hat{\dist}_{k-1}^x)
  }_{\substack{\text{current prediction}\\\text{error}}}
  +
  \underbrace{
  \ell_{\mathrm{reg},k-1}^x(\hat{\dist}_{k-1}^x)
  }_{\substack{\text{disturbance}\\\text{regularization}}}
  \notag\\
  &\quad+
  \underbrace{
  \ell_{\mathrm{adm},k-1}^x(\hat{\dist}_{k-1}^x)
  }_{\substack{\text{physical}\\\text{admissibility}}}
  \notag\\
  \ell_{\mathrm{pred},k-1}^x(\hat{\dist}_{k-1}^x)
  &=
  \left\|
    W_x\vect{e}_{k}^{x}(\hat{\dist}_{k-1}^x)
  \right\|^2,
  \notag\\
  \ell_{\mathrm{reg},k-1}^x(\hat{\dist}_{k-1}^x)
  &=
  \left\|
    W_d^x\hat{\dist}_{k-1}^{x}
  \right\|_2^2,
  \notag\\
  \ell_{\mathrm{adm},k-1}^x(\hat{\dist}_{k-1}^x)
  &=
  R_{\mathrm{adm}}^x\!\left(
    \hat{\dist}_{k-1}^x,
    \state_{k-1},
    \action_{k-1},
    \state_{k}^{\fb}
  \right).
  \notag
\end{align}

The prediction term measures the feedback mismatch, with $W_x$ setting the relative importance of the feedback coordinates. The regularization term discourages unnecessarily large disturbance estimates, with $W_d^x$ weighting the disturbance coordinates. The penalty function $R_{\mathrm{adm}}^x$ penalizes violations of known physical constraints, as illustrated in \figref{fig:mdob_three_term_system}. For example, a candidate disturbance exceeding a known magnitude bound incurs a penalty.

When Eq.~\eqref{eq:state_effect_argmin} can be solved analytically, the disturbance is estimated directly. For sampled or black-box rollout models, SDOB instead draws $m_x$ candidate disturbances for the completed transition,
\begin{equation}
  \hat{\dist}_{k-1,r}^x
  \sim
  \rho_{k-1}^x,
  \qquad
  r=1,\ldots,m_x .
  \label{eq:state_effect_disturbance_samples}
\end{equation}
Here $\rho_{k-1}^x$ denotes the proposal probability distribution used to sample candidate state-effect disturbances. It may be centered on the previous disturbance estimate. Each candidate is inserted into the loss in Eq.~\eqref{eq:state_effect_argmin}. Candidates with lower losses receive larger exponential weights:
\begin{align}
  \ell_{k-1,r}^x
  &=
  \mathcal L_{k-1}^x(\hat{\dist}_{k-1,r}^x),
  \label{eq:state_effect_sample_loss}
  \\
  \omega_{k-1,r}^x
  &=
  \frac{\exp(-\ell_{k-1,r}^x/\lambda_d)}
       {\sum_{q=1}^{m_x} \exp(-\ell_{k-1,q}^x/\lambda_d)} .
  \label{eq:state_effect_sample_weight}
\end{align}
The normalized exponential weights define an empirical Gibbs distribution \citep{theodorou2010_path_integral,williams2017_mppi_tro}, with $\lambda_d>0$ controlling its concentration. The point estimate used in the experiments is its weighted mean,
\begin{equation}
  \hat{\dist}_{k-1}^x
  =
  \sum_{r=1}^{m_x}\omega_{k-1,r}^x\hat{\dist}_{k-1,r}^x .
  \label{eq:weighted_state_effect_disturbance}
\end{equation}
The weighted mean provides a practical point estimate by combining the sampled disturbances according to their losses, without additional model evaluations once the candidates have been scored. This estimate need not coincide with a minimizer of Eq.~\eqref{eq:state_effect_argmin}. The same fitting objective could also be addressed by selecting the candidate with the lowest loss or by iteratively refitting the sampling distribution to an elite set using the CEM \citep{kobilarov2012_cross_entropy_planning,pinneri2020_cem}. These are alternative sampling-based solution strategies. For simplicity of presentation, we use the weighted mean throughout this paper for sampled state-effect estimation. The estimate becomes available at step $k$, remains constant over the next prediction horizon, and is computed using only forward rollouts.

\subsection{Cost Disturbance Estimation}

The cost channel estimates how environmental changes affect planning costs while the task objective remains fixed. Following the design intuition of compression as intelligence, it uses local cost queries to retain task-relevant information without reconstructing the full environment or tracking every object. For example, a moving obstacle shifts the surrounding high-cost region. SDOB estimates this local motion of the cost field from queries before and after an environment update and uses it to predict future costs, without requiring an explicit model of obstacle motion.

We denote the local velocity of the cost field by
$\dist^c(\query,t)$ and refer to it as the cost disturbance. Its admissible
values form $\mathcal D_c$. This velocity describes motion in query space
and need not equal the physical velocity of an individual obstacle.

For a differentiable cost field, the relation between its local motion and
its change over time is
\begin{equation}
  \frac{\partial \pcost(\query,t)}{\partial t}
  +
  \left[\dist^c(\query,t)\right]^{\!\top}
  \nabla_{\query}\pcost(\query,t)
  =
  s^c(\query,t).
  \label{eq:cost_transport_balance}
\end{equation}
The source term $s^c(\query,t)$ accounts for cost changes that the local
motion does not explain, such as a newly appearing obstacle. When this term
is negligible, the cost remains approximately unchanged along the motion of
the field, following the same idea as brightness constancy in optical flow
\citep{horn1981_optical_flow}.

SDOB assumes that this approximation holds between consecutive cost updates.
Let $\Delta t$ be the time between these updates. The displacement over
$[k-1,k]$ is approximately $\Delta t\,\dist_{k-1}^c(\query)$, giving
\begin{equation}
  \pcost_k(\query)
  \approx
  \pcost_{k-1}
  \left(
    \query-\Delta t\,\dist_{k-1}^c(\query)
  \right).
  \label{eq:cost_transport_discretization}
\end{equation}
In other words, the current cost at $\query$ is matched to the previous
cost at a position shifted backward by the estimated displacement.
SDOB evaluates this relation directly through cost queries, without computing
cost gradients. It can therefore also be applied to nonsmooth costs and
binary collision queries. Changes that cannot be explained by the shift
remain in the matching residual.

\begin{figure*}[!t]
\centering
\includegraphics[width=0.98\textwidth]{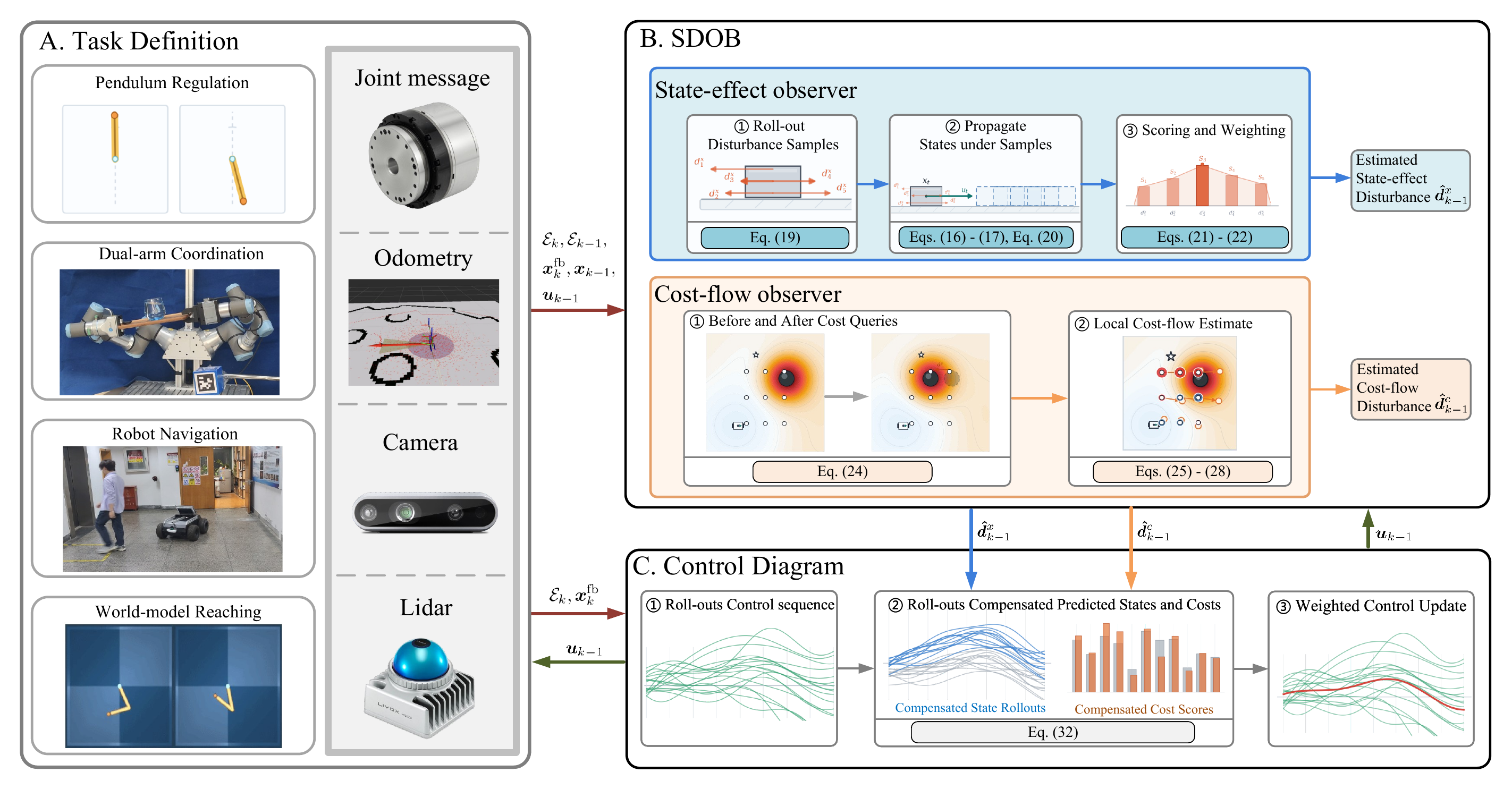}

\caption{Schematic of the SDOB computation flow. Feedback updates the state-effect and cost disturbance estimates for subsequent rollout and cost compensation.}
\label{fig:mdob_control_flow}
\end{figure*}

To estimate the local motion, SDOB selects $N_c$ query states,
$\mathcal Q_k^c=\{\query_k^{(i)}\}_{i=1}^{N_c}$, at step $k$.
\figref{fig:cost_flow_estimation_pipeline} illustrates the process with
$N_c=9$, with the selected backward shifts shown in the third panel. A cost change
at one point is generally insufficient to determine a motion vector. Each candidate velocity $\hat{\dist}_{k-1,i}^c$ is therefore
tested over a neighborhood $\mathcal N_i$ around query state
$\query_k^{(i)}$, assuming approximately uniform motion within that
neighborhood. For each neighbor $j$, the candidate shifts the query position
in the previous cost field to predict the current cost:
\begin{equation}
  \hat{\pcost}_{k|k-1}^{c,(i,j)}
  \left(\hat{\dist}_{k-1,i}^c\right)
  =
  \pcost_{k-1}
  \left(
    \query_k^{(j)}
    -
    \Delta t\,\hat{\dist}_{k-1,i}^c
  \right),
  \label{eq:cost_disturbance_prediction}
\end{equation}
The difference between the current queried cost and this prediction is
\begin{equation}
  e_{k,i,j}^{c}
  \left(\hat{\dist}_{k-1,i}^c\right)
  =
  \pcost_k(\query_k^{(j)})
  -
  \hat{\pcost}_{k|k-1}^{c,(i,j)}
  \left(\hat{\dist}_{k-1,i}^c\right).
  \label{eq:cost_flow_residual}
\end{equation}
The local estimate minimizes the weighted matching error across the
neighborhood, together with a penalty on the velocity magnitude:
\begin{align}
  \hat{\dist}_{k-1,i}^{c,\star}
  &\in
  \argmin_{\hat{\dist}_{k-1,i}^c\in\mathcal D_c}
  \mathcal L_{k-1,i}^c(\hat{\dist}_{k-1,i}^c),
  \label{eq:cost_disturbance_argmin}\\
  \mathcal L_{k-1,i}^c(\hat{\dist}_{k-1,i}^c)
  &=
  \underbrace{
  \sum_{j\in\mathcal N_i}
  W_{ij}^c
  \left[e_{k,i,j}^{c}(\hat{\dist}_{k-1,i}^c)\right]^2
  }_{\text{local cost matching}}
  \notag\\
  &\quad+
  \underbrace{
  \left\|W_d^c\hat{\dist}_{k-1,i}^c\right\|_2^2
  }_{\text{transport regularization}}.
  \notag
\end{align}
The nonnegative weights $W_{ij}^c$ set each neighbor's contribution and sum to one over
$\mathcal N_i$. The matrix $W_d^c$ controls the penalty on large velocities.
SDOB tests a finite set of sampled or gridded velocities in $\mathcal D_c$
and retains the candidate with the lowest loss at each query state.

Rollouts may query positions outside the selected set. SDOB interpolates the
cost-field velocity at these positions by averaging nearby estimates, weighted
by distance and confidence.
The resulting kernel interpolation
\citep{nadaraya1964_regression,kang2016_kernel_approximation} is
\begin{align}
  Z_c(\query)
  &=
  \sum_{j=1}^{N_c}
  W_c^{(j)}
  K\!\left(\|\query-\query_k^{(j)}\|/\sigma_c\right),
  \notag\\
  \alpha_i(\query)
  &=
  \begin{cases}
  \dfrac{
    W_c^{(i)}K\!\left(\|\query-\query_k^{(i)}\|/\sigma_c\right)
  }{Z_c(\query)}, & Z_c(\query)>0,\\[0.7em]
  0, & Z_c(\query)=0,
  \end{cases}
  \notag\\
  \hat{\dist}_{k-1}^c(\query)
  &=
  \sum_{i=1}^{N_c}
  \alpha_i(\query)\hat{\dist}_{k-1,i}^c .
  \label{eq:cost_disturbance_interpolation}
\end{align}
Here $W_c^{(i)}\geq0$ is the confidence of estimate $i$, and $K$ is a
nonnegative kernel with spatial scale $\sigma_c>0$. The factor
$Z_c(\query)$ normalizes the weights. If no local estimate contributes,
$Z_c(\query)=0$ and the velocity at $\query$ is set to zero.

Under the slowly time-varying disturbance assumption, SDOB holds the estimated
velocity field fixed over the prediction horizon. To predict the cost at
$\query$ after $h$ steps, it traces that position backward through the
estimated field for $h$ steps, then queries the current cost at the resulting
position. Starting from $\query_h^{[0]}=\query$, the backward steps are
\begin{equation}
  \query_h^{[r+1]}
  =
  \query_h^{[r]}
  -
  \Delta t\,
  \hat{\dist}_{k-1}^c\!\left(\query_h^{[r]}\right),
  \quad r=0,\ldots,h-1.
  \label{eq:cost_query_backtrace}
\end{equation}
The predicted cost is then
\begin{equation}
  \widehat{\pcost}_{k+h|k}^{\,c}(\query)
  =
  \pcost_k\!\left(\query_h^{[h]}\right),
  \quad h=1,\ldots,H.
  \label{eq:cost_query_transport}
\end{equation}
The minus sign in Eq.~\eqref{eq:cost_query_backtrace} traces $\query$
backward along the estimated motion to the corresponding location in
the current cost field. Equation~\eqref{eq:cost_query_transport} then
uses the cost at this location to predict the cost of future rollouts.

\begin{remark}[Local cost queries and limitations]
Cost motion cannot be identified in constant-cost regions and is ambiguous
along straight obstacle boundaries. Multiple local motions or obstacle
appearance and disappearance may violate the transport assumption, while
incomplete observations near sensing or map boundaries can introduce
spurious cost changes. In practice, restricting $\query$ to a region around
the robot or its predicted trajectory helps keep query neighborhoods and
shifted positions within the reliably observed area. This reduces errors
near sensing or map boundaries without resolving local motion ambiguities.
\end{remark}

\subsection{Using SDOB in Control}

\figref{fig:mdob_control_flow} and Algorithm~\ref{alg:mdob_control_step}
summarize how SDOB enters the control update. At step $k$,
$\hat{\dist}_{k-1}^x$ compensates the state rollouts, while the cost disturbance
constructs the horizon query function $\widehat{\pcost}_{k+h|k}^{\,c}$ through
Eqs.~\eqref{eq:cost_query_backtrace} and \eqref{eq:cost_query_transport}. Let

\begin{equation}
  U_k
  =
  \left(
    \action_{k|k},
    \ldots,
    \action_{k+H-1|k}
  \right)
  \label{eq:horizon_control_sequence}
\end{equation}
be a horizon control sequence. The controller solves
\begin{align}
  U_k^\star
  &=
  \argmin_{U_k}
  S_k(U_k),
  \label{eq:mdob_control_optimization}
  \\
  S_k(U_k)
  &=
  \underbrace{
  \Phi_{\mathrm{cost},k}(U_k)
  }_{\substack{\text{cost channel}\\\text{term}}}
  +
  \underbrace{
  \Phi_{\mathrm{other},k}(U_k)
  }_{\substack{\text{other}\\\text{constraint cost}}}
  +
  \notag\\
  &\quad
  \underbrace{
  \Phi_{g,k}(U_k)
  }_{\substack{\text{goal}\\\text{cost}}}
  +
  \underbrace{
  \Phi_{u,k}(U_k)
  }_{\substack{\text{u}\\\text{smooth}}},
  \notag
  \\
  \Phi_{\mathrm{cost},k}(U_k)
  &=
  \sum_{h=1}^{H}
  \widehat{\pcost}_{k+h|k}^{\,c}
  \left(\query_{k+h|k}\right),
  \notag\\
  \Phi_{\mathrm{other},k}(U_k)
  &=
  \sum_{h=0}^{H-1}
  \ell_k^{\mathrm{con}}
  \left(\state_{k+h|k},\action_{k+h|k}\right),
  \notag\\
  \Phi_{g,k}(U_k)
  &=
  F_c^f(\state_{k+H|k},\state_g),
  \notag\\
  \Phi_{u,k}(U_k)
  &=
  \sum_{h=0}^{H-1}
  \ell_k^{u,\mathrm{sm}}
  \left(\Delta\action_{k+h|k}\right),
  \notag
  \\
  \text{s.t.}\quad
  \state_{k|k}
  &=
  \state_k^{\fb},
  \notag\\
  \state_{k+h+1|k}
  &=
  F_x^d(\state_{k+h|k},\action_{k+h|k},\hat{\dist}_{k-1}^x).
  \notag
\end{align}
The objective contains four terms. The cost channel term $\Phi_{\mathrm{cost},k}$ evaluates costs exposed through the cost-query interface, such as collision cost, at each predicted task query state using the transported query function. The term $\Phi_{\mathrm{other},k}$ collects constraints not handled through this interface, such as state or input limits, and is not modified by cost disturbance compensation. The goal term $\Phi_{g,k}$ measures the predicted goal error, while $\Phi_{u,k}$ penalizes changes between adjacent inputs to promote smooth control commands.

\begin{algorithm}[t]
\small
\caption{SDOB control update}
\label{alg:mdob_control_step}
\KwIn{Transition $(\state_{k-1},\action_{k-1},\state_k^{\fb})$; queries $\pcost_{k-1},\pcost_k$; $F_x^d$, $\Pi_q$, and $H$}
\KwOut{$\hat{\dist}_{k-1}^x$, $\hat{\dist}_{k-1}^c(\cdot)$, and $\action_k$}
\tcp{State-effect estimation}
Sample $\hat{\dist}_{k-1,r}^x\sim\rho_{k-1}^x$, $r=1,\ldots,m_x$\;
\For{$r=1,\ldots,m_x$}{
  Predict $\hat{\state}_{k|k-1}^{(r)}\gets F_x^d(\state_{k-1},\action_{k-1},\hat{\dist}_{k-1,r}^x)$\;
  Compute $\vect e_k^{x,(r)}\gets\state_k^{\fb}-\hat{\state}_{k|k-1}^{(r)}$\;
  $\ell_{k-1,r}^x\gets\mathcal L_{k-1}^x(\hat{\dist}_{k-1,r}^x)$ via \eqref{eq:state_effect_argmin}\;
}
Normalize weights $\omega_{k-1,r}^x$ via \eqref{eq:state_effect_sample_weight}\;
Set $\hat{\dist}_{k-1}^x\gets\sum_{r=1}^{m_x}\omega_{k-1,r}^x\hat{\dist}_{k-1,r}^x$\;
\tcp{Cost disturbance estimation}
Select $\mathcal Q_k^c$ and neighborhoods $\mathcal N_i$\;
\For{$i=1,\ldots,N_c$}{
  Evaluate candidate cost residuals over $\mathcal N_i$ via \eqref{eq:cost_flow_residual}\;
  Select the minimum-loss velocity $\hat{\dist}_{k-1,i}^c$ via \eqref{eq:cost_disturbance_argmin}\;
}
Interpolate $\hat{\dist}_{k-1}^c(\cdot)$ using \eqref{eq:cost_disturbance_interpolation}\;
\tcp{Compensated control}
\ForEach{candidate control sequence $U_k$}{
  Initialize $\state_{k|k}\gets\state_k^{\fb}$\;
  \For{$h=0,\ldots,H-1$}{
    Predict $\state_{k+h+1|k}\gets F_x^d(\state_{k+h|k},\action_{k+h|k},\hat{\dist}_{k-1}^x)$\;
    Extract $\query_{k+h+1|k}\gets\Pi_q(\state_{k+h+1|k})$\;
    Backtrace $\query_{k+h+1|k}$ and query its cost via \eqref{eq:cost_query_backtrace}--\eqref{eq:cost_query_transport}\;
  }
  Evaluate $S_k(U_k)$ via \eqref{eq:mdob_control_optimization}\;
}
Solve \eqref{eq:mdob_control_optimization} for $U_k^\star$\;
Apply $\action_k=\action_{k|k}^{\star}$\;
\end{algorithm}

\section{Simulation and Experiments}

We evaluate SDOB through five simulation studies and three robot studies spanning analytical and learned prediction models. The comparisons assess the task-level performance of complete controller configurations incorporating SDOB, with task-specific disturbance representations and prediction interfaces. Each study addresses a particular use of feedback correction, and its conclusions are interpreted within the corresponding task conditions and execution protocol.

\begin{table*}[!t]
\centering
\caption{Core SDOB and controller parameters used in simulation.}
\label{tab:simulation_parameters}
{\footnotesize
\setlength{\tabcolsep}{5pt}
\begin{tabular}{@{}llccccc@{}}
\toprule
Task & \shortstack{SDOB\\channel} & \shortstack{SDOB state\\samples $m_x$} & \shortstack{SDOB query\\states $N_c$} & \shortstack{Controller rollout\\samples} & \shortstack{Prediction\\horizon $H$} & \shortstack{Environment\\step} \\
\midrule
Inverted pendulum & State effect & 256 & -- & 1024 & 15 & 0.050 s \\
Eight-obstacle navigation & State and cost & 256 & 332 & 500 & 30 & 0.050 s \\
Latent Reacher & State effect & 256 & -- & 300 & 5 & 0.020 s \\
Rocket Collect & Cost & -- & 289 & 32 & 12 & 0.033 s \\
Landing & State and cost & 256 & 121 & 32 & 24 & 0.033 s \\
\bottomrule
\end{tabular}}
\end{table*}

Within each simulation task, methods share the nominal model, nominal task objective, action limits, initial conditions, evaluation seeds, prediction horizon, and nominal rollout budget. Observer, robustness, and cost-correction mechanisms form part of the evaluated controller configurations; their effects on rollout and cost evaluation are described in the task-specific setups below. Repeated runs use paired task or planner seeds, and method-specific settings remain fixed. SDOB is compared with the corresponding nominal controller and, when available, DOB, Tube-MPPI, and RMPPI. Reported planning time measures wall-clock computation and includes controller-specific observer and robustness computations. The simulation tasks and the real-robot controller computations use an NVIDIA GeForce RTX 4080 Laptop GPU. Table~\ref{tab:simulation_parameters} lists the core simulation settings; the remaining observer and controller parameters are provided in the open-source configuration files.

\begin{figure*}[!t]
\centering
\includegraphics[width=0.98\textwidth]{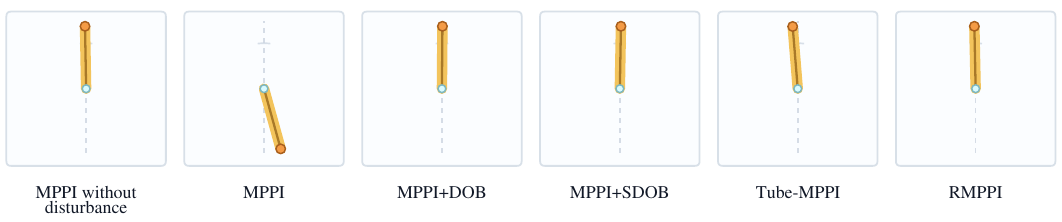}
\caption{Representative inverted-pendulum snapshots under sinusoidal
torque bias. MPPI+SDOB effectively compensates for the torque disturbance, achieving disturbance-rejection performance comparable to the conventional MPPI+DOB.}
\label{fig:pendulum_snapshot}
\par\vspace{10pt}
\begin{minipage}{\textwidth}
\captionsetup*{type=table}
\centering
\caption{Inverted-pendulum simulation results.}
\label{tab:pendulum_multirun}
{\footnotesize
\setlength{\tabcolsep}{4.5pt}
\begin{tabular}{@{}lcccccc@{}}
\toprule
Method & Success & \shortstack{Tail $|\theta|$\\(rad)} & \shortstack{Settling\\time (s)} & \shortstack{Disturbance\\RMSE} & \shortstack{Control\\TV} & \shortstack{P95 planning\\time (ms)} \\
\midrule
MPPI (no disturbance) & 20/20 & $0.008 \pm 0.004$ & $0.897 \pm 0.203$ & -- & $0.125 \pm 0.009$ & $8.274 \pm 0.541$ \\
\midrule
MPPI & 0/20 & $1.108 \pm 0.006$ & -- & -- & $0.172 \pm 0.010$ & $8.024 \pm 0.462$ \\
MPPI+DOB & \textbf{20/20} & $0.027 \pm 0.004$ & $2.413 \pm 0.142$ & $0.018 \pm 0.001$ & $\boldsymbol{0.127 \pm 0.007}$ & $\boldsymbol{8.018 \pm 0.405}$ \\
MPPI+SDOB & \textbf{20/20} & $0.026 \pm 0.004$ & $\boldsymbol{2.402 \pm 0.157}$ & $\boldsymbol{0.014 \pm 0.001}$ & $\boldsymbol{0.127 \pm 0.007}$ & $8.232 \pm 0.426$ \\
Tube-MPPI & 19/20 & $0.081 \pm 0.008$ & $2.668 \pm 0.662$ & -- & $0.162 \pm 0.007$ & $16.914 \pm 1.192$ \\
RMPPI & \textbf{20/20} & $\boldsymbol{0.021 \pm 0.006}$ & $2.423 \pm 0.227$ & -- & $0.178 \pm 0.010$ & $19.628 \pm 1.143$ \\
\bottomrule
\end{tabular}}
\end{minipage}
\end{figure*}

The analytical-model studies examine state-effect compensation and navigation under combined state and cost disturbances. The learned-model studies examine recovery from an imposed action bias, correction of environment-dependent costs, and initial feedback calibration for planning with a frozen predictor.

Closed-loop task performance is assessed using success rate together with task-specific tracking and safety metrics. Command smoothness is quantified by control total variation (TV)
\begin{equation}
  \operatorname{TV}(\action)
  =
  \frac{1}{T-1}
  \sum_{k=0}^{T-2}
  \left\lVert \action_{k+1}-\action_k \right\rVert_2
  \label{eq:control_total_variation}
\end{equation}
where a lower value indicates smaller changes between consecutive commands. Computation time is summarized by the 95th percentile (P95) of per-step planning time. Tabulated values with $\pm$ report the mean and standard deviation across trials. Success and collision entries report trial counts.

\subsection{Simulation Results}

\paragraph{1) Inverted-pendulum state-effect task:}
This study evaluates state-effect compensation with an analytical nominal model and compares SDOB with a conventional DOB and robust MPPI variants. The task is an inverted pendulum subject to a sinusoidal torque disturbance. \figref{fig:pendulum_snapshot} shows representative snapshots under the common disturbance, and Table~\ref{tab:pendulum_multirun} summarizes 20 seeds.

Except for the disturbance-free nominal reference, all methods are evaluated under the same sinusoidal torque disturbance $d(t)=1.2\sin(1.2t)\,\mathrm{N\,m}$. Success requires $|\theta|<0.12$ rad for at least the final 20\% of the rollout. The uncompensated MPPI controller fails in all 20 trials, whereas MPPI+DOB, MPPI+SDOB, and RMPPI succeed in all trials and Tube-MPPI succeeds in 19. MPPI+DOB and MPPI+SDOB have similar tail error and control variation, but MPPI+SDOB reduces the disturbance estimation RMSE from $0.018$ to $0.014$. RMPPI achieves a smaller tail error at more than twice the P95 planning time of MPPI+SDOB.

\begin{figure*}[!t]
\centering
\includegraphics[width=\textwidth]{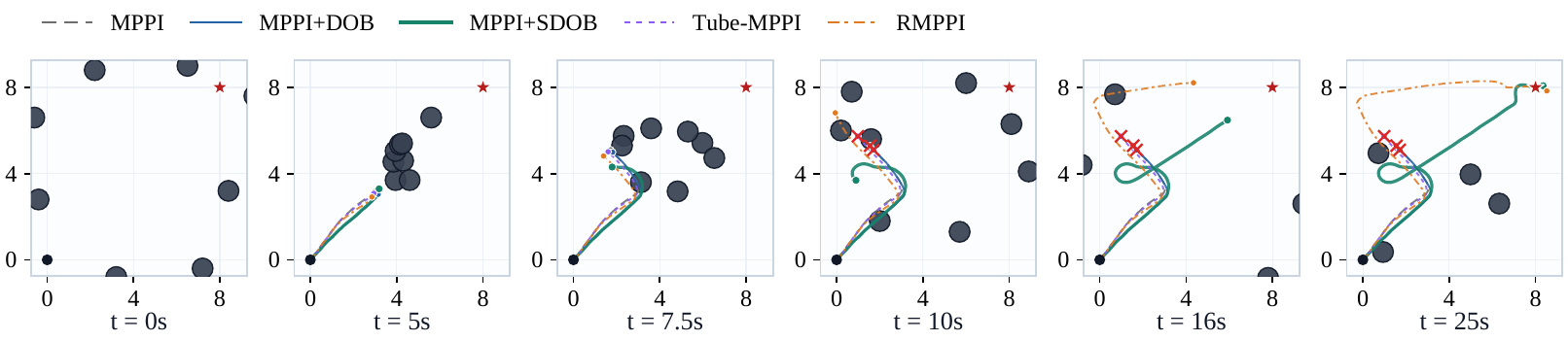}
\caption{Snapshots comparing navigation methods in an eight-obstacle environment with state-effect and cost disturbances. MPPI+SDOB successfully avoids collisions while achieving a shorter navigation path than the comparison methods.}
\label{fig:nav_8obs_frames}
\par\vspace{10pt}
\begin{minipage}{\textwidth}
\captionsetup*{type=table}
\centering
\caption{Eight-obstacle navigation simulation results.}
\label{tab:navigation_multirun}
{\footnotesize
\setlength{\tabcolsep}{2pt}
\begin{tabular}{@{}lccccccc@{}}
\toprule
Method & Success & Collision & \shortstack{Min. clearance\\(m)} & \shortstack{Goal time\\(s)} & \shortstack{Path\\ratio} & \shortstack{Control\\TV} & \shortstack{P95 planning\\time (ms)} \\
\midrule
MPPI & 0/20 & 20/20 & $-0.006 \pm 0.004$ & -- & -- & $0.697 \pm 0.053$ & $\boldsymbol{7.368 \pm 0.708}$ \\
MPPI+DOB & 0/20 & 20/20 & $-0.009 \pm 0.004$ & -- & -- & $0.689 \pm 0.051$ & $8.736 \pm 1.410$ \\
MPPI+SDOB & \textbf{20/20} & \textbf{0/20} & $\boldsymbol{0.129 \pm 0.034}$ & $21.323 \pm 2.846$ & $\boldsymbol{1.484 \pm 0.171}$ & $0.554 \pm 0.034$ & $20.770 \pm 1.661$ \\
Tube-MPPI & 0/20 & 20/20 & $-0.002 \pm 0.002$ & -- & -- & $0.674 \pm 0.046$ & $7.771 \pm 0.889$ \\
RMPPI & \textbf{20/20} & \textbf{0/20} & $0.125 \pm 0.010$ & $\boldsymbol{21.225 \pm 0.026}$ & $1.573 \pm 0.003$ & $\boldsymbol{0.518 \pm 0.025}$ & $18.764 \pm 1.146$ \\
\bottomrule
\end{tabular}}
\end{minipage}
\end{figure*}

\paragraph{2) Eight-obstacle navigation with state-effect and cost disturbances:}
This study evaluates the complete SDOB configuration in an eight-obstacle navigation task where both state and cost disturbances affect planning. A constant input bias changes the robot motion, while moving obstacles change the queried costs. The obstacles approach from several directions and may overlap. All controllers use the same obstacle margin. \figref{fig:nav_8obs_frames} shows six instants, and Table~\ref{tab:navigation_multirun} reports 20 seeds.

\begin{figure*}[!t]
\centering
\includegraphics[width=\textwidth]{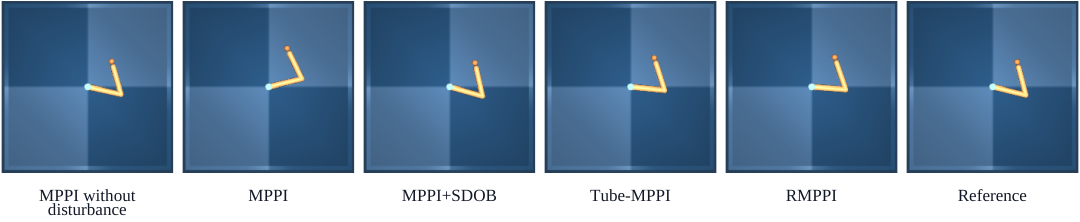}
\caption{Final Reacher observations from representative evaluations, where MPPI+SDOB achieves a smaller reaching error.}
\label{fig:reacher_final_states}
\par\vspace{10pt}
\begin{minipage}{\textwidth}
\captionsetup*{type=table}
\centering
\caption{ Reacher simulation results.}
\label{tab:reacher_result}
{\footnotesize
\setlength{\tabcolsep}{3.8pt}
\begin{tabular}{@{}lcccccc@{}}
\toprule
Method & Success & \shortstack{Tail mean\\error (rad)} & \shortstack{Final\\error (rad)} & \shortstack{Disturbance\\RMSE} & \shortstack{Control\\TV} & \shortstack{P95 planning\\time (ms)} \\
\midrule
MPPI (no disturbance) & 20/20 & $0.017 \pm 0.004$ & $0.016 \pm 0.007$ & -- & $0.104 \pm 0.011$ & $438.956 \pm 41.523$ \\
\midrule
MPPI & 0/20 & $0.427 \pm 0.038$ & $0.431 \pm 0.038$ & -- & $\boldsymbol{0.125 \pm 0.016}$ & $\boldsymbol{456.264 \pm 69.320}$ \\
MPPI+SDOB & \textbf{19/20} & $\boldsymbol{0.030 \pm 0.010}$ & $\boldsymbol{0.031 \pm 0.012}$ & $\boldsymbol{0.056 \pm 0.011}$ & $0.126 \pm 0.020$ & $502.569 \pm 5.313$ \\
Tube-MPPI & 0/20 & $0.185 \pm 0.062$ & $0.220 \pm 0.105$ & -- & $0.163 \pm 0.051$ & $492.350 \pm 38.858$ \\
RMPPI & 0/20 & $0.176 \pm 0.044$ & $0.208 \pm 0.109$ & -- & $0.152 \pm 0.043$ & $492.228 \pm 41.545$ \\
\bottomrule
\end{tabular}}
\end{minipage}
\end{figure*}

Each run has a 30 s limit, and success means reaching the $0.35$ m goal region before collision. The path ratio is the traveled path length divided by the straight-line distance from the start to the goal, so a lower value indicates a more direct route. Goal time and path ratio use successful trials only; negative clearance denotes overlap. MPPI, MPPI+DOB, and Tube-MPPI collide in all trials, whereas MPPI+SDOB and RMPPI succeed in all trials. MPPI+SDOB has a shorter mean path ratio than RMPPI with similar clearance, while its local query update gives the largest planning time.

\paragraph{3) Latent world-model Reacher task:}
This experiment examines whether SDOB can use prediction errors in the model's latent space to compensate for an imposed action bias and recover performance on tasks already solvable by the frozen model and nominal planner. The second state-effect study uses the pretrained Reacher checkpoint released with LeWorldModel (LeWM) \citep{maes2026_leworldmodel}. The controller encodes current and goal images, samples actions with MPPI, and scores latent rollouts. A constant action bias $[0.30,-0.20]$ is applied during execution but hidden from the planner. SDOB estimates an equivalent action bias from the latent endpoint mismatch. The controller uses images and latent states; joint angles are used only for evaluation. Because the model does not expose the local linear dynamics required by LQR, Tube-MPPI and RMPPI use MPPI as their auxiliary controller.

\begin{figure*}[!t]
\centering

\begin{minipage}[t]{0.48\textwidth}
\captionsetup{type=figure}
\centering
\includegraphics[width=\linewidth]{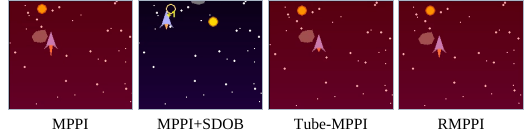}
\caption{Final Rocket collect observations from representative evaluations, where MPPI+SDOB reduces collisions and improves task success.}
\label{fig:mmbench2_rocket_collect}
\end{minipage}\hfill
\begin{minipage}[t]{0.48\textwidth}
\captionsetup{type=figure}
\centering
\includegraphics[width=\linewidth]{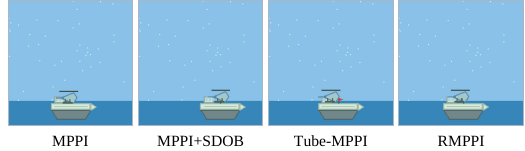}
\caption{Final Landing observations from representative evaluations,where MPPI+SDOB achieves a smaller touchdown error.}
\label{fig:mmbench2_landing}
\end{minipage}

\par\vspace{10pt}
\begin{minipage}{\textwidth}
\centering
\captionsetup*{type=table}
\caption{MMBench2 simulation results over 20 seeds. Touchdown center errors use trials that reach the deck.}
\label{tab:mmbench2_results}
{\footnotesize
\setlength{\tabcolsep}{3pt}
\begin{tabular}{@{}lcccc@{\hspace{10pt}}cccc@{}}
\toprule
\multirow{2}{*}{Method} & \multicolumn{4}{c}{Rocket Collect} & \multicolumn{4}{c}{Visual Landing} \\
\cmidrule(lr){2-5}\cmidrule(lr){6-9}
& Success & Collision & \shortstack{Control\\TV} & \shortstack{P95 planning\\time (s)} & Success & \shortstack{Touchdown center\\error (px)} & \shortstack{Control\\TV} & \shortstack{P95 planning\\time (s)} \\
\midrule
MPPI & 16/20 & 4/20 & $\boldsymbol{0.409 \pm 0.071}$ & $\boldsymbol{2.116 \pm 0.093}$  & \textbf{20/20} & $20.49 \pm 10.60$ & $0.955 \pm 0.144$ & $\boldsymbol{12.131 \pm 1.057}$  \\
MPPI+SDOB & \textbf{19/20} & \textbf{0/20} & $0.421 \pm 0.071$ & $2.219 \pm 0.201$  & \textbf{20/20} & $\boldsymbol{7.38 \pm 6.90}$ & $0.835 \pm 0.203$ & $12.454 \pm 1.072$  \\
Tube-MPPI & 16/20 & 4/20 & $0.518 \pm 0.071$ & $2.491 \pm 0.206$  & \textbf{20/20} & $14.54 \pm 4.70$ & $\boldsymbol{0.764 \pm 0.162}$ & $12.556 \pm 0.717$  \\
RMPPI & 15/20 & 4/20 & $0.537 \pm 0.130$ & $2.310 \pm 0.171$  & \textbf{20/20} & $16.61 \pm 4.60$ & $0.795 \pm 0.122$ & $12.670 \pm 0.986$  \\
\bottomrule
\end{tabular}}
\end{minipage}
\end{figure*}

The pretrained LeWM model has prediction limitations that can prevent reliable task completion even without an additional action disturbance. To evaluate compensation for the added disturbance on tasks that the nominal model and planner can already solve, we select 20 tasks for which disturbance-free MPPI has a final-20-step mean error no greater than $0.025$ rad, then evaluate every disturbed method on the same tasks and planner seeds. Success requires this tail error to be below $0.05$ rad. As shown in \figref{fig:reacher_final_states} and Table~\ref{tab:reacher_result}, MPPI+SDOB succeeds in 19 trials and lowers the mean tail error to $0.030$ rad, versus $0.427$, $0.185$, and $0.176$ rad for MPPI, Tube-MPPI, and RMPPI. Within this nominally solvable task subset, SDOB recovered successful performance in 19 of 20 trials under the imposed action bias.

\begin{figure*}[!t]
\centering
\includegraphics[width=\textwidth]{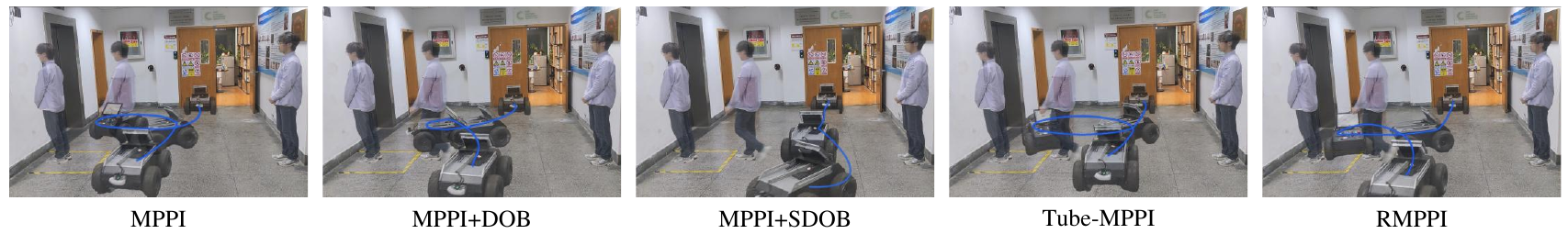}
\caption{Representative real-car navigation snapshots, where MPPI+SDOB achieves faster collision-free navigation.}
\label{fig:real_car_navigation}
\par\vspace{10pt}
\includegraphics[width=\textwidth]{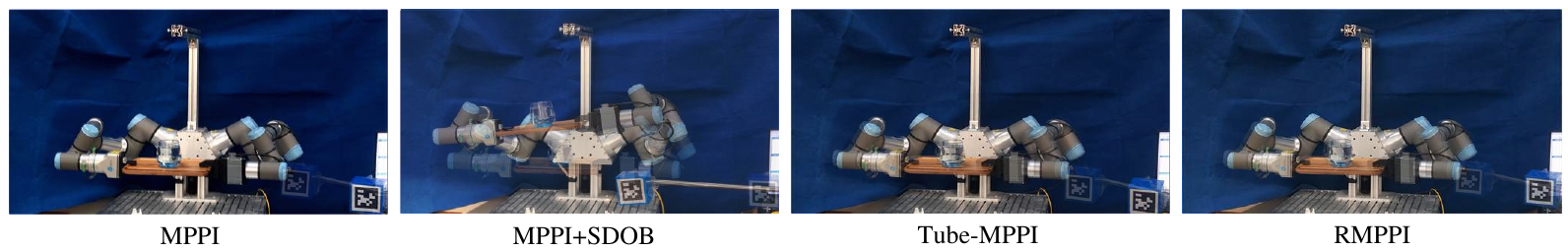}
\caption{Representative dual-arm obstacle-avoidance snapshots, where SDOB responds earlier to the moving obstacle and successfully avoids collision.}
\label{fig:dual_arm_obstacle_avoidance}

\end{figure*}

\paragraph{4) Rocket Collect cost-disturbance task:}
This study evaluates the complete cost-compensated controller under faster obstacle motion, using reconstructed cost maps to augment world-model planning scores.
The rocket must collect a coin while avoiding moving asteroids. For control, the available feedback consists of the current RGB image and the rocket's position and velocity. The released MMBench2~\citep{hansen2026_mmbench2} world model uses the image and candidate action sequences to predict rewards and collision-risk scores through its reward head. A nominal motion model predicts rocket positions along each candidate path, and the coin detected in the image supplies the goal. The controller combines the learned scores with the distance to the coin and penalties on control effort and variation. For SDOB, the decoder reconstructs each current image, from which asteroid pixels are extracted to define an obstacle-distance cost map. The cost observer needs two successive maps queried at the same $17\times17$ grid points to estimate how the cost pattern moves. The controller then shifts the current map according to this estimate and adds a penalty where the predicted motion raises the cost along a candidate path. Thus, the reward head supplies the basic planning scores, while the reconstructed cost maps supply the SDOB correction. We increase asteroid speed to $1.5$ times its nominal value and replan every two environment steps. MPPI, MPPI+SDOB, Tube-MPPI, and RMPPI are tested on the same 20 random seeds with the same world model and rollout budget. Success requires collecting the coin without collision within 140 steps. The controller has no access to simulator-provided obstacle states or collision geometry.

\figref{fig:mmbench2_rocket_collect} and Table~\ref{tab:mmbench2_results} show that MPPI+SDOB succeeds in 19/20 trials with no collisions, compared with 16/20 for MPPI and Tube-MPPI and 15/20 for RMPPI. Its runs have slightly higher control variation than MPPI, while planning time remains comparable. For these evaluated seeds, the cost-compensated configuration produced fewer collisions under the specified increase in asteroid speed.

\begin{table}[t]
\centering
\caption{Completion times in the representative real-car runs.}
\label{tab:real_car_completion_time}
{\small
\begin{tabular}{@{}lc@{}}
\toprule
Method & Completion time (s) \\
\midrule
MPPI & 12.13 \\
MPPI+DOB & 14.04 \\
\textbf{MPPI+SDOB} & \textbf{9.29} \\
Tube-MPPI & 12.07 \\
RMPPI & 13.10 \\
\bottomrule
\end{tabular}}
\end{table}

\paragraph{5) Visual Landing task:}

This study evaluates touchdown alignment when input bias and deck motion affect the landing task.
The helicopter must land on a moving deck. For control, the inputs are the current RGB image and a fixed landing reference image. Helicopter and deck positions are detected in the current image, and successive helicopter positions provide a velocity estimate. The released MMBench2 world model scores candidate actions using predicted rewards and similarity to the reference image in its feature space. A nominal motion model also predicts helicopter positions and velocities, allowing the controller to evaluate distance to the detected deck and touchdown safety. SDOB uses two sets of feedback. The state observer takes recent image-based helicopter positions, the executed control inputs, and the nominal motion model to estimate an input bias that explains the observed motion. This estimate is included in both the world-model and motion-model predictions when evaluating new actions. The cost observer queries the landing cost defined by the detected deck at the same $11\times11$ grid points across successive observations. It estimates the motion of this cost map, which the controller uses to move its future landing targets. We add an unknown horizontal input bias of $0.16$ and increase deck speed to $1.6$ times nominal. The same four methods are tested on the same 20 random seeds with a common world model, replanning every four environment steps within a 96-step limit. First deck contact counts as success; touchdown center error measures landing accuracy. Simulator state and deck velocity are used only for evaluation.

As shown in \figref{fig:mmbench2_landing} and Table~\ref{tab:mmbench2_results}, all methods reach the deck in 20/20 trials. MPPI+SDOB reduces mean touchdown center error to $7.38$ pixels, compared with $20.49$ for MPPI, $14.54$ for Tube-MPPI, and $16.61$ for RMPPI. Tube-MPPI has lower control variation, indicating that SDOB's main benefit in this task is improved touchdown alignment.

\begin{remark}
In the world-model simulations, planning is already slow without disturbance estimation: baseline MPPI has mean P95 planning times of $0.46$~s, $2.12$~s, and $12.13$~s for Reacher, Rocket Collect, and Landing, respectively. These exceed the corresponding simulation-time control intervals, and disturbance estimation adds further computation. In these three tasks, the simulator is therefore paused during planning, including disturbance estimation, and advances only after computation finishes. Environment-step durations and replanning intervals refer to simulation time; reported planning times measure wall-clock computation. Reducing the combined computation time of planning and disturbance estimation is therefore an urgent priority for real-time feedback control with world models.
\end{remark}

\begin{figure*}[t]
\centering
\includegraphics[width=0.85\textwidth]{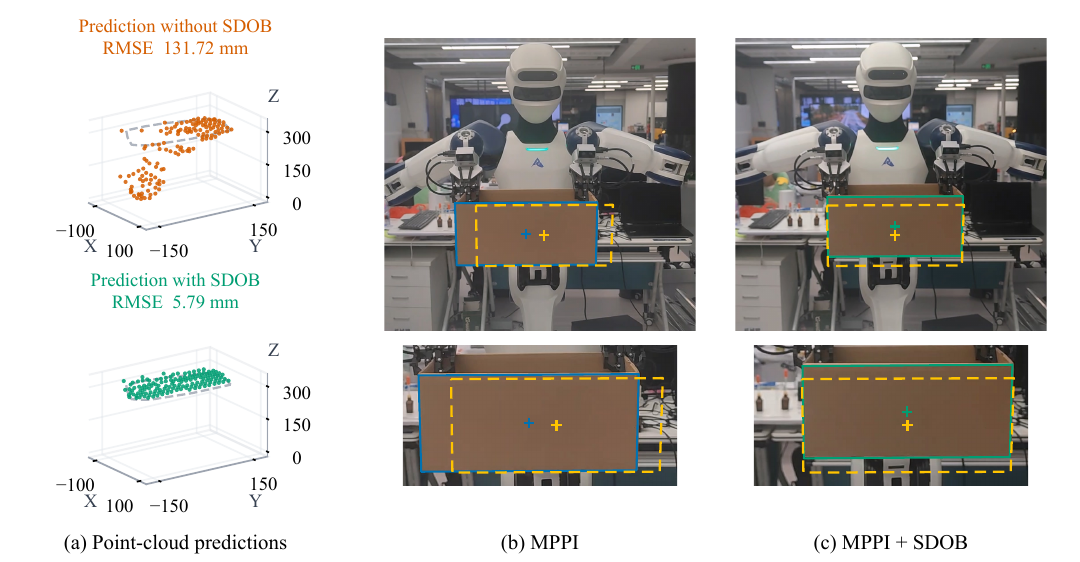}
\caption{PointWorld box-lifting results. (a) Point-cloud motion
predictions without and with SDOB.
(b) Real-robot results with MPPI.
(c) Real-robot results with MPPI+SDOB, showing a smaller box-lifting error.}
\label{fig:box_lift_combined}

\end{figure*}

\subsection{Experimental Results}

Three robot studies illustrate SDOB with different prediction interfaces. Vehicle and dual-arm obstacle avoidance provide representative closed-loop demonstrations with state-space models. In the PointWorld~\citep{huang2026_pointworld} box-lifting study, SDOB uses initial feedback to correct model predictions for planning. The resulting plan is then executed without replanning, and task performance is compared over 20 paired trials.

\paragraph{1) Real-car obstacle-avoidance navigation task:}
Vehicle motion is compared with nominal rollouts, while a moving person changes the perceived cost field. No separate person detection is performed; SDOB estimates the resulting cost-field motion from temporal changes in the cost queries. No ground-truth disturbance label is available, so the trial is assessed by route choice, collision avoidance, and completion time. In the representative runs in \figref{fig:real_car_navigation}, MPPI+SDOB follows the shortest route and finishes in $9.29$ s, as listed in Table~\ref{tab:real_car_completion_time}. 

\paragraph{2) Dual-arm obstacle-avoidance task:}
The minimum collision distance is set to $0$ cm, so no additional safety margin is imposed. The obstacle carries an AprilTag \citep{olson2011_apriltag} to update its current pose in the collision model, but no independent measurement of its Cartesian velocity or future trajectory is provided to the controller. SDOB instead estimates the obstacle motion from temporal changes in the collision-cost queries. In the representative trials in \figref{fig:dual_arm_obstacle_avoidance}, MPPI+SDOB begins avoidance earlier and avoids the obstacle, while the shown comparison runs collide.

\begin{table}[t]
\centering
\caption{PointWorld box-lifting results over 20 paired trials.}
\label{tab:box_lift_20_pairs}
{\small
\setlength{\tabcolsep}{3pt}
\begin{tabular}{@{}lcc@{}}
\toprule
Metric & MPPI & MPPI+SDOB \\
\midrule
Final task RMSE (mm)& $55.26 \pm 14.99$ & $22.61 \pm 12.58$ \\
Final lift (mm) & $287.39 \pm 24.91$ & $306.60 \pm 12.73$ \\

\bottomrule
\end{tabular}}
\end{table}

\paragraph{3) PointWorld box lifting with initial feedback calibration:}

A dual-arm robot lifts a grasped box by 300~mm using a frozen PointWorld
model. This experiment examines feedback
calibration for planning with the frozen predictor. Forward inference is too
slow for real-time replanning in the current implementation, so the study
compares MPPI and MPPI+SDOB using a single plan, with no replanning or observer
updates during execution. An initial 50~mm lift
provides RGB-D feedback to calibrate the state-effect correction. Alongside
the PointWorld prediction, a rigid-body prediction assumes that the grasped
box follows the translation and rotation of the two hands.
This rigid-body prediction specifies the task-specific correction prior. SDOB samples
a scalar correction coefficient between zero and one and weights the candidates
according to how closely their corrected predictions match the observed
box motion during this initial lift. For each subsequent candidate action,
the estimated coefficient scales the difference between its rigid-body and
PointWorld predictions. This correction is added to the predicted box points
at each horizon step. The estimated coefficient remains fixed, while the correction varies
with the candidate hand trajectory. MPPI then evaluates the corrected box
motion against the lifting target to select an action sequence, with the
PointWorld model weights kept frozen. As shown in
\figref{fig:box_lift_combined}, the comparison includes point-cloud motion
predictions and representative real-robot lifting results. The results over
20 trials are summarized in Table~\ref{tab:box_lift_20_pairs}.
The feedback-calibrated configuration achieved a lower mean final task
RMSE than the uncorrected PointWorld planner over these trials.

\FloatBarrier
\section{Conclusion and Discussion}
This paper presented SDOB for disturbance estimation and compensation through state-rollout and cost-query interfaces. The state channel estimates corrections to predicted motion, while the cost channel estimates local changes in the cost field. Simulation and robot studies demonstrated improved task performance across analytical models, simulators, and learned world models.

Current world models have two limitations for disturbance-aware control.
First, planning with these models requires repeated forward evaluations
and already takes too long for high-frequency closed-loop control.
Additional model evaluations for disturbance estimation further increase
this computational burden. Second, their latent representations often
lack a clear physical interpretation, making it difficult to represent
disturbances and relate latent prediction errors to physical effects.
Our world-model demonstrations are therefore limited to relatively simple
tasks with task-specific disturbance representations. Future work will
explore more efficient joint planning and disturbance estimation, as well
as models trained with explicit disturbance inputs to support compensation
in more complex tasks.

\section*{Acknowledgments}
The authors used generative AI to assist with code writing and debugging. The authors take responsibility for the final implementation and the reported results.

\section*{Code Availability}
The code release includes source code, configurations, and simulation
scripts for mobile-robot navigation, the inverted pendulum, and LeWM Reacher
compensation, available at
\url{https://sampling-based-dob.github.io/}.

{\urlstyle{rm}
\interlinepenalty=10000
\bibliographystyle{SdobSageH}
\bibliography{references}
}

\end{document}